\documentclass[runningheads]{llncs}
\usepackage[T1]{fontenc}
\usepackage{amsmath}
\usepackage{booktabs}
\usepackage{graphicx}
\begin{document}
\title{Who Gets Missed in the Tail?\\
Thresholded Subgroup Underdiagnosis in Long-Tailed Chest X-ray Classification}
\titlerunning{Thresholded Subgroup Underdiagnosis in CXR}
\author{
% Ha-Hieu Pham\inst{1,2,3}\orcidID{0009-0003-8938-2745} \and
Ha-Hieu Pham\inst{1,2,3} \and
Hai-Dang Nguyen\inst{3,6} \and
Dang P. M. Cao\inst{6} \and
Thanh-Huy Nguyen\inst{4} \and
Min Xu\inst{4} \and
Trung-Nghia Le\inst{1,2} \and
Ulas Bagci\inst{5} \and
Huy-Hieu Pham\inst{3,6} 
}

\authorrunning{H.-H. Pham et al.}

\institute{
University of Science, Ho Chi Minh City, Vietnam \and
Vietnam National University, Ho Chi Minh City, Vietnam \and
VinUni-Illinois Smart Health Center, VinUniversity, Hanoi, Vietnam \and
Carnegie Mellon University, Pittsburgh, PA, USA \and
Northwestern University, Chicago, IL, USA \and
College of Engineering \& Computer Science, VinUniversity, Hanoi, Vietnam 
% \email{hieu.ph2@vinuni.edu.vn}
}

\maketitle              % typeset the header of the contribution
\begin{abstract}
In chest X-ray (CXR) classification, acceptable ranking performance can still leave rare-positive patients below threshold, especially within subgroups.
We study this pre-deployment fairness problem as an audit question: after a long-tailed multi-label CXR model is converted from scores into decisions, who is missed?
Across VinDr-CXR and MIMIC-CXR/CXR-LT, we use a diagnostic ladder to separate class-level long-tail losses, subgroup-aware weighting, group robustness, and threshold selection.
On VinDr-CXR, group-tail weighting followed by tail-aware thresholding reduces tail FNR from 0.665 to 0.269, sex worst-group FNR from 0.705 to 0.157, and age worst-group FNR from 0.822 to 0.133, while macro-mAP increases from 0.611 to 0.635.
On MIMIC-CXR/CXR-LT, the same score-to-threshold comparison reduces tail FNR from 0.866 to 0.741 and lowers worst-group FNR across sex, age, race, and insurance; residual missed-positive rates nevertheless remain high.
Paired bootstrap contrasts on VinDr support the thresholded FNR reductions, and GroupDRO reference runs indicate that aggregate group robustness alone does not remove rare subgroup misses in this setting.
The study supports a narrow audit claim: rare-label fairness in CXR depends jointly on the finding, subgroup, and operating threshold, not on label frequency or ranking metrics alone.
\end{abstract}
\section{Introduction}
\label{sec:intro}

A CXR triage model ultimately acts through thresholds: a case is surfaced for review only if its score crosses an operating point.
This matters in the long tail, where rare findings have limited positive support and subgroup-specific positives may be rarer still.
A model can therefore have acceptable macro-mAP while leaving rare-positive cases in a particular subgroup below threshold.
The fairness question is not only whether rare labels are difficult, but whether thresholded decisions concentrate missed positives in specific class-subgroup cells.

We study long-tailed CXR fairness as a deployment-oriented measurement problem.
The audit unit is a positive rare-label case within a subgroup after a class-specific threshold has been applied.
We evaluate this unit on VinDr-CXR, where subgroup metadata are limited to DICOM-derived sex and age, and on MIMIC-CXR/CXR-LT~\cite{Johnson2019,Johnson2023,LIN2025103739}, where EHR-linked metadata also support race and insurance analyses.
Rather than proposing a deployable diagnostic system or a general fairness remedy, we use model variants as diagnostic probes.
The goal is to localize where missed positives enter the pipeline: class imbalance, subgroup support, group-robust training, or the score-to-decision threshold.

Existing long-tail methods such as effective-number reweighting~\cite{Cui2019} and asymmetric loss~\cite{Tal2021} act primarily at the class level.
Fairness methods such as equal opportunity~\cite{Hardt2016} and GroupDRO~\cite{Sagawa2020} typically aggregate over groups, while medical-imaging fairness studies have documented demographic and EHR-derived bias in clinical prediction pipelines~\cite{Gianfrancesco2020,Larrazabal2020,SeyyedKalantari2021}.
Recent CXR work further shows that locally fair models may not remain fair under distribution shift~\cite{Yang2024,Dong2026}, and AI diagnostic reporting guidance emphasizes bias, fairness, and applicability~\cite{Sounderajah2025}.
Our setting combines these concerns: the relevant error is not a rare class alone or a demographic group alone, but a rare $(\text{class}, \text{subgroup})$ positive evaluated after thresholding.

\noindent\textbf{Contributions.}
We make four contributions.
First, we define thresholded class-subgroup underdiagnosis as an audit unit for long-tailed CXR fairness.
Second, we use a diagnostic ladder that separates class-level losses, subgroup-aware weighting, group robustness, and threshold selection.
Third, we report subgroup-conditioned FNR, positive support, missed positives per 100 positives, and bootstrap contrasts to keep small-cell evidence interpretable.
Fourth, across two CXR cohorts, we show that ranking metrics and aggregate group robustness do not determine who is missed after thresholding; the operating point materially changes observed tail harm.

\section{Study Design}
\label{sec:method}

\noindent\textbf{Cohort-specific observability.}
Fairness claims are bounded by the subgroup metadata and label definitions available in each cohort.
We therefore keep each dataset in its native label space, subgroup definitions, and train/validation/test split, and define tail classes from training-set frequency within that cohort.
This design avoids uncertain label mappings and makes the supported fairness questions explicit.
MIMIC-CXR/CXR-LT is a second within-cohort instantiation of the audit rather than a mapped-label transfer experiment.

\noindent\textbf{Data.}
VinDr-CXR~\cite{Nguyen2022} contains 15K training images and 15 labels, with a 1,500-image test split used here.
Subgroups are sex and age extracted from DICOM headers; missing or non-informative values are retained for accounting and excluded from headline sex comparisons when appropriate.
Race and insurance are not available in VinDr, so we do not project US EHR subgroup definitions onto this dataset.
MIMIC-CXR/CXR-LT \cite{Johnson2019,Johnson2023,LIN2025103739} uses the MIMIC-CXR image collection with CXR-LT labels and a 78{,}946-image test split; subgroup axes are sex, age group, race, and insurance from available metadata.
An axis is reported only when metadata are available and clinically interpretable.

\begin{table}[t]
\centering
\scriptsize
\setlength{\tabcolsep}{4.8pt}
\renewcommand{\arraystretch}{1.12}
\caption{Cohort observability and tail support. Tail classes are defined within each cohort from training-set frequency; subgroup axes are reported only when present in the cohort metadata.}
\label{tab:cohort_observability}
\resizebox{\linewidth}{!}{%
\begin{tabular}{lcccl}
\toprule
\textbf{Cohort} & \textbf{Test studies} & \textbf{Labels / tail classes} & \textbf{Tail positives} & \textbf{Reported subgroup axes} \\
\midrule
VinDr-CXR & 1{,}500 & 15 / 5 & 56 total; 3--19 per tail class & Sex, age group from DICOM headers; race and insurance unavailable \\
MIMIC-CXR/CXR-LT & 78{,}946 & 40 / 13 & 2{,}889 total; 48--454 per tail class & Sex, age group, race, and insurance from EHR-linked metadata \\
\bottomrule
\end{tabular}
}
\end{table}

\noindent\textbf{Diagnostic ladder.}
All probes use a ConvNeXt-Tiny backbone with a dataset-specific linear head, so the ladder changes the training objective or thresholding rule rather than the architecture.
M1 is BCE, M2 uses asymmetric loss (ASL)~\cite{Tal2021}, and M3 adds effective-number class weighting~\cite{Cui2019}.
M4 is a GroupDRO reference baseline~\cite{Sagawa2020}.
M5 multiplies ASL by a clipped class-subgroup weight based on training counts.
M6 reuses M5 scores and changes only tail-class thresholds.
M1--M5 use validation $F_1$ thresholds; for M6, head/medium classes keep $F_1$ thresholds, while tail-class thresholds maximize worst-group recall subject to a minimum precision floor $\tau=0.05$.
Thus M5 and M6 have identical ranked scores, ECE, and Brier scores; any M5--M6 FNR change is threshold-mediated.

\noindent\textbf{Endpoints and uncertainty.}
Ranking is summarized by macro-mAP and tail-mAP.
Thresholded underdiagnosis is summarized by tail FNR and worst-group tail FNR for each available subgroup axis.
ECE and Brier score describe calibration, and missed positives per 100 true positives translate FNR into a clinically interpretable unit.
Tail-class FNRs on VinDr rest on only 3--19 positives per class, so point estimates are low-resolution.
We report independent bootstrap 95\% CIs using 200 test-row resamples and paired bootstrap contrasts using 1,000 shared resamples for method comparisons.

\section{Experiments and Results}
\label{sec:exp}

\subsection{VinDr-CXR Diagnostic Ladder}

\begin{table}[h]
\centering
\scriptsize
\setlength{\tabcolsep}{4.4pt}
\renewcommand{\arraystretch}{1.10}
\caption{VinDr-CXR test results with bootstrap 95\% CIs (200 resamples over the 1{,}500-image test split). M1--M5 use F1-optimal thresholds; M6 reuses M5 scores with tail-aware thresholds.}
\label{tab:vindr_full}
\resizebox{\linewidth}{!}{%
\begin{tabular}{lcccccc}
\toprule
\textbf{Metric} & \textbf{M1 BCE} & \textbf{M2 ASL} & \textbf{M3 CB-ASL} & \textbf{M4 GroupDRO} & \textbf{M5 GT-ASL} & \textbf{M6 +TA} \\
\midrule
Macro-mAP   & 0.611 [0.56,0.68] & 0.616 [0.57,0.69] & 0.588 [0.54,0.65] & 0.648 [0.59,0.72] & 0.635 [0.60,0.70] & 0.635 [0.60,0.70] \\
Tail-mAP    & 0.484 [0.35,0.66] & 0.521 [0.40,0.69] & 0.485 [0.35,0.64] & 0.564 [0.43,0.74] & 0.558 [0.44,0.72] & 0.558 [0.44,0.72] \\
Tail FNR    & 0.665 [0.51,0.82] & 0.551 [0.38,0.69] & 0.614 [0.45,0.76] & 0.540 [0.37,0.71] & 0.471 [0.33,0.64] & 0.269 [0.08,0.40] \\
Sex wg-FNR  & 0.705 [0.53,0.90] & 0.557 [0.39,0.76] & 0.613 [0.47,0.81] & 0.616 [0.45,0.82] & 0.507 [0.35,0.71] & 0.157 [0.03,0.38] \\
Age wg-FNR  & 0.822 [0.50,1.00] & 0.549 [0.27,0.89] & 0.756 [0.42,0.94] & 0.554 [0.37,0.86] & 0.633 [0.33,0.83] & 0.133 [0.00,0.50] \\
ECE         & 0.012 [0.01,0.02] & 0.103 [0.10,0.11] & 0.137 [0.13,0.14] & 0.044 [0.04,0.05] & 0.056 [0.06,0.06] & 0.056 [0.06,0.06] \\
Brier       & 0.020 [0.02,0.02] & 0.037 [0.03,0.04] & 0.048 [0.05,0.05] & 0.023 [0.02,0.03] & 0.025 [0.02,0.03] & 0.025 [0.02,0.03] \\
\bottomrule
\end{tabular}
}
\end{table}

\noindent Table~\ref{tab:vindr_full} reports the primary test-set analysis for VinDr-CXR (1,500-image test set).
All thresholds for M1--M5 are F1-optimal per class and tuned on the validation split; M6 uses the same M5 checkpoint but replaces the threshold objective on tail classes with the tail-aware calibration rule in Sec.~\ref{sec:method}.
This separates score-model changes (M1--M5) from the effect of moving the operating point without retraining (M6).

\subsection{Large-Cohort Audit on MIMIC-CXR/CXR-LT}

\noindent Table~\ref{tab:mimic_compact} reports the same ladder on MIMIC-CXR/CXR-LT.
The pattern is directionally consistent with VinDr but smaller in magnitude: M6 lowers tail FNR and worst-group FNR across all reported subgroup axes relative to M5, while macro-mAP and tail-mAP are unchanged because M6 reuses M5 scores.
Absolute MIMIC tail FNR remains high, so thresholding exposes and reduces part of the miss rate rather than solving the tail.
Compared with the M5 F1 operating point, M6 reduces MIMIC tail FNR from 0.866 to 0.741, sex worst-group FNR from 0.912 to 0.795, age worst-group FNR from 0.983 to 0.873, race worst-group FNR from 0.945 to 0.880, and insurance worst-group FNR from 0.902 to 0.819.
The most consistent cross-cohort pattern is that moving the operating point on an unchanged score model changes thresholded subgroup FNR more than class-level long-tail variants do.
The MIMIC operating sweep shows that this effect is not unique to M5: at the same $\tau=0.05$ precision floor, tail-aware thresholds reduce tail FNR for BCE (0.859 to 0.773), ASL (0.864 to 0.731), CB-ASL (0.851 to 0.732), and GroupDRO (0.888 to 0.806).
MIMIC's richer subgroup axes extend the audit beyond sex and age, while remaining cohort-specific observability rather than a universal fairness taxonomy.

\begin{table}[h]
\centering
\scriptsize
\setlength{\tabcolsep}{4.8pt}
\renewcommand{\arraystretch}{1.12}
\caption{MIMIC-CXR/CXR-LT test results (78{,}946 images). M1--M5 use F1-optimal thresholds; M6 reuses M5 scores with tail-aware thresholds. Lower FNR is better.}
\label{tab:mimic_compact}
\resizebox{\linewidth}{!}{%
\begin{tabular}{lccccccc}
\toprule
\textbf{Method} & \multicolumn{2}{c}{\textbf{Ranking}} & \multicolumn{5}{c}{\textbf{Thresholded underdiagnosis}} \\
\cmidrule(lr){2-3}\cmidrule(lr){4-8}
& \textbf{Macro-mAP} & \textbf{Tail-mAP} & \textbf{Tail FNR} & \textbf{Sex wg-FNR} & \textbf{Age wg-FNR} & \textbf{Race wg-FNR} & \textbf{Ins. wg-FNR} \\
\midrule
M1 BCE      & 0.224 & 0.067 & 0.859 & 0.908 & 0.949 & 0.960 & 0.905 \\
M2 ASL      & 0.225 & 0.067 & 0.864 & 0.928 & 0.979 & 0.965 & 0.919 \\
M3 CB-ASL   & 0.225 & 0.069 & 0.851 & 0.912 & 0.982 & 0.964 & 0.893 \\
M4 GroupDRO & 0.215 & 0.060 & 0.888 & 0.945 & 0.977 & 0.984 & 0.940 \\
M5 GT-ASL   & 0.223 & 0.067 & 0.866 & 0.912 & 0.983 & 0.945 & 0.902 \\
M6 +TA      & 0.223 & 0.067 & 0.741 & 0.795 & 0.873 & 0.880 & 0.819 \\
\bottomrule
\end{tabular}
}
\end{table}

\begin{figure}[t]
\centering
\includegraphics[width=0.96\linewidth]{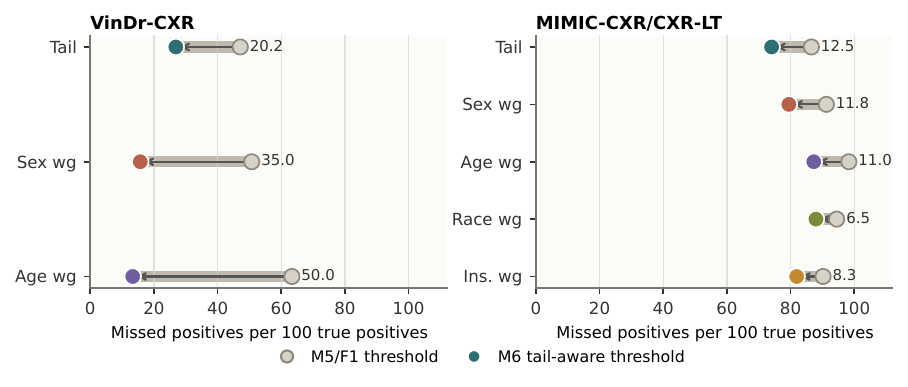}
\caption{M5 to M6 threshold effect across cohorts. Grey markers show M5 with F1 thresholds; colored markers show the same score model after tail-aware thresholding. Numbers are absolute reductions in missed positives per 100 true positives.}
\label{fig:threshold_effect}
\end{figure}

\subsubsection{What the ladder diagnoses.}

\noindent The ladder exposes failure modes that macro-mAP alone would hide.
On VinDr, the BCE baseline reaches macro-mAP 0.611 but misses roughly two-thirds of tail positives after thresholding, with higher FNR in the worst sex and age groups.
ASL improves average tail behavior without reliably protecting the worst group, and CB-ASL leaves substantial age worst-group FNR.
Because M6 keeps the M5 scores and changes only the tail threshold objective, the M5--M6 reduction isolates the score-to-action step.
From M1 to M6, VinDr tail positives missed per 100 fall from 66.5 to 26.9, worst-sex-subgroup misses from 70.5 to 15.7, and worst-age-subgroup misses from 82.2 to 13.3.
M6's independent bootstrap interval for tail FNR, [0.08,0.40], does not overlap the M1 interval, [0.51,0.82], supporting the headline reduction despite the small test set.
MIMIC shows the same qualitative threshold effect at larger scale, but with much higher residual missed-positive rates, reinforcing the measurement interpretation rather than a claim of solved mitigation.

\subsubsection{Tail-class support.}

\noindent The aggregate M6 gain is not uniformly distributed across diseases.
It is driven by Calcification ($n_+=19$, FNR 0.895 to 0.158), ILD ($n_+=16$, 0.375 to 0.000), and Consolidation ($n_+=11$, 0.818 to 0.091).
Pneumothorax ($n_+=7$) changes little, and Atelectasis ($n_+=3$) is below the resolution at which a 1{,}500-image test set can support stable disease-level conclusions.
This support check keeps the disease-level interpretation tied to the available positive counts.

\subsubsection{Paired contrasts and reference baseline.}
\label{sec:exp_paired}

\noindent Paired bootstrap resampling uses shared test-row indices across methods, with contrasts chosen to isolate the contribution of each component: M6 vs M1 for the full ladder, M6 vs M5 for thresholding, and M5 vs M3 for subgroup-aware training over class-only reweighting.
Table~\ref{tab:paired_bootstrap} reports the mean paired difference $\mu(A)-\mu(B)$, its 95\% interval, and the empirical probability that method $A$ has a lower FNR than method $B$.
For FNR metrics, negative values are better.

\begin{table}[h]
\centering
\scriptsize
\setlength{\tabcolsep}{6.0pt}
\renewcommand{\arraystretch}{1.12}
\caption{Paired bootstrap contrasts on VinDr-CXR ($n_{\text{boot}}=1{,}000$ except age rows with one degenerate resample omitted). Negative FNR differences favor the first method.}
\label{tab:paired_bootstrap}
\begin{tabular}{llccc}
\toprule
\textbf{Contrast} & \textbf{Metric} & \textbf{Mean diff.} & \textbf{95\% CI} & \textbf{$P(A < B)$} \\
\midrule
M6 vs M1 & Tail FNR & -0.402 & [-0.517,-0.302] & 1.000 \\
M6 vs M1 & Sex wg-FNR & -0.531 & [-0.764,-0.331] & 1.000 \\
M6 vs M1 & Age wg-FNR & -0.614 & [-0.944,-0.286] & 1.000 \\
M6 vs M5 & Tail FNR & -0.210 & [-0.353,-0.041] & 0.992 \\
M6 vs M5 & Sex wg-FNR & -0.353 & [-0.569,-0.134] & 0.998 \\
M6 vs M5 & Age wg-FNR & -0.433 & [-0.757,-0.110] & 0.995 \\
M5 vs M3 & Tail FNR & -0.141 & [-0.292,-0.041] & 1.000 \\
M5 vs M3 & Sex wg-FNR & -0.111 & [-0.243,-0.021] & 0.984 \\
M5 vs M3 & Age wg-FNR & -0.106 & [-0.304,0.000] & 0.882 \\
\bottomrule
\end{tabular}
\end{table}

The paired results support the main interpretation.
M6 improves all three FNR endpoints over M1 and over the identical-score M5 operating point, indicating that the threshold objective accounts for a substantial part of the reduction.
M5 also improves tail FNR over M3, suggesting that subgroup-aware weighting contributes before calibration, although the age worst-group contrast is weaker.

M4 GroupDRO is included in Tables~\ref{tab:vindr_full} and~\ref{tab:mimic_compact} as a group-robustness reference rather than a direct ladder rung.
It does not match M6 on thresholded tail or worst-group FNR.
On MIMIC, applying the same tail-aware rule to GroupDRO lowers tail FNR from 0.888 to 0.806 at $\tau=0.05$, but remains worse than M6 on tail FNR and all reported worst-group FNR endpoints.
This reference does not argue against group robustness in general.
It shows that aggregate group robustness and thresholded rare-positive recall are distinct audit dimensions.

\subsubsection{Tail-aware ablation, alert burden, and seed stability.}
\label{sec:exp_operating}

\noindent The M5/M6 comparison isolates the effect of thresholding on one score model.
Applying the same tail-aware rule to M1--M3 and M5 reduces tail and worst-group FNR across score models, indicating that the phenomenon is not specific to one favorable checkpoint.
The relevant claim is narrower: thresholded subgroup underdiagnosis is a deployment-stage failure mode that ranking metrics do not expose.

The ablation also exposes the workload cost of reducing misses. For M5 on VinDr, the default F1 operating point predicts 4.9 tail-positive labels per 100 studies at tail FNR 0.471, while the $\tau=0.05$ tail-aware point predicts 38.9 per 100 studies at tail FNR 0.269.
A stricter floor, $\tau=0.10$, reduces the alert rate to 18.2 per 100 studies while preserving a lower tail FNR (0.313) than the F1 operating point.
On MIMIC, the corresponding M5 comparison is less dramatic but still clinically material: F1 thresholding predicts 6.4 tail-positive labels per 100 studies at tail FNR 0.866; $\tau=0.05$ predicts 19.3 per 100 studies at tail FNR 0.741; and $\tau=0.10$ predicts 7.1 per 100 studies at tail FNR 0.811.
The operating point is therefore a clinical policy choice, not a universal parameter, and Table~\ref{tab:mimic_compact} should be interpreted as evidence about thresholded underdiagnosis rather than as a workflow recommendation.

\begin{figure}[t]
\centering
\includegraphics[width=0.96\linewidth]{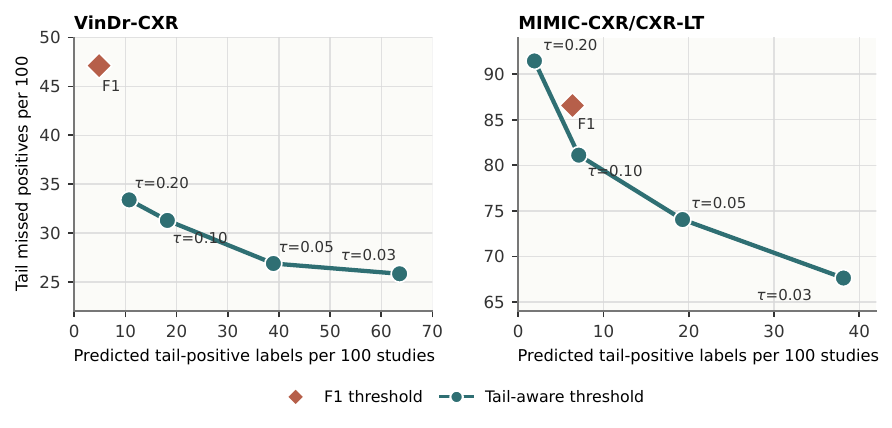}
\caption{Operating-point trade-off for the M5 score model in both cohorts. Moving from the F1 threshold to tail-aware thresholds lowers missed tail positives but increases predicted tail-positive labels per 100 studies; the choice of $\tau$ is therefore a workflow policy rather than a universal setting.}
\label{fig:operating_tradeoff}
\end{figure}

Finally, the VinDr three-seed aggregate indicates that the qualitative ranking is not driven by a single initialization: M6 has the lowest mean tail FNR (0.249$\pm$0.051), sex worst-group FNR (0.173$\pm$0.018), and age worst-group FNR (0.089$\pm$0.038), while M1--M5 remain substantially higher on thresholded endpoints.
This seed analysis is still VinDr-only and should not be read as cross-cohort stability.

\section{Interpretation}
\label{sec:discuss}

\noindent The diagnostic ladder is intended as a measurement instrument, not a leaderboard.
It follows rare findings from support, to score ranking, to thresholded decisions, and then asks which subgroup remains missed.
Across the two cohorts, the long tail is better understood as a collection of thresholded class-subgroup cells whose reliability depends on support, metadata coverage, and operating policy.

\noindent\textbf{The tail is a mixture, not a regime.}
The VinDr support check illustrates why a single tail average is insufficient.
M6 sharply improves Calcification, ILD, and Consolidation, changes Pneumothorax little, and cannot stabilize Atelectasis with only three test positives.
Rare findings therefore differ in whether they are threshold-sensitive, sample-limited, or simply too sparse for stable disease-level claims.

\noindent\textbf{The patient at risk is defined after conditioning.}
Class imbalance alone does not identify who is missed.
A finding may be rare overall, but the safety-relevant error is a positive case in a subgroup after thresholding.
This is why class-level remedies can improve an average tail endpoint while leaving a worst-off subgroup with high FNR.
In VinDr, ASL improves average tail behavior but changes which sex is worst-off; in MIMIC, class-level variants leave tail and worst-group FNR close to baseline.
Because M6 lowers reported worst-group FNR endpoints without changing ranking metrics, the audit unit must include the class, subgroup, and thresholded decision.

\noindent\textbf{Thresholding is the score-to-action boundary.}
M5 and M6 have identical ranked scores, yet their missed-positive rates differ once the tail threshold objective changes from F1 to worst-group recall with a precision floor.
For CXR triage or screening, threshold documentation is therefore part of fairness reporting rather than an implementation detail.

\noindent\textbf{Fairness is bounded by observability and threshold policy.}
VinDr supports a focused sex/age analysis but not race or insurance; MIMIC supports additional EHR-derived axes but inherits their measurement limits.
Lower tail thresholds reduce missed positives only if the resulting review burden remains usable, so metadata coverage, positive support, missed positives per 100, and threshold policy must be reported together.

\vspace{0.3em}
\noindent\textbf{Clinical relevance and limitations.}
For screening, triage, and worklist prioritization, false negatives are the most direct safety concern because they remove true positives from downstream attention.
Reporting missed positives per 100 true positives makes this failure mode easier to interpret than an isolated FNR.
Group-tail reweighting is operationally simple because it requires only training-set counts for each class-subgroup pair, but here it is a diagnostic probe rather than a deployment recommendation.
Tail-aware thresholding is clinically meaningful only when paired with precision or review-burden constraints.
The value of $\tau$ should be selected with clinical stakeholders and validated under the intended workflow rather than treated as universal.

The evidence has important bounds. The two cohorts support the same measurement concern but do not establish deployment readiness or architecture-general fairness.
VinDr's 1.5K-image test split contains only 3--19 positives per tail class, and many subgroup-tail cells have $\le 5$ positives; we therefore report bootstrap CIs and positive counts and interpret disease-specific estimates as low-resolution signals.
VinDr DICOM headers also provide incomplete subgroup metadata: only 49\% of test images have informative sex and \texttt{PatientAge} is missing in 69\%.
MIMIC-CXR/CXR-LT provides richer subgroup axes, but EHR-derived race and insurance remain imperfect social and administrative proxies rather than causal attributes.
The study is observational, uses a single ConvNeXt-Tiny backbone, and does not support causal, deployment-ready, or architecture-general claims.

\section{Conclusion}
\label{sec:conc}

We present a deployment-oriented fairness evaluation for long-tailed CXR classification that follows rare findings from ranking metrics to thresholded subgroup decisions.
Across VinDr-CXR and MIMIC-CXR/CXR-LT, class-level long-tail variants and a GroupDRO reference do not fully determine which rare positives remain below threshold.
The most consistent lesson is methodological: rare-label fairness requires reporting support, subgroup-conditioned false negatives, uncertainty, workload trade-offs, and the threshold policy that converts scores into action.
Threshold-aware auditing can reduce observed subgroup misses in this setting, but high residual FNRs and cohort-specific metadata limits show why it should be treated as a fairness measurement tool rather than a deployment-ready remedy.

\bibliographystyle{splncs04}
\bibliography{mybibliography}

@article{Johnson2019,
  author = {Johnson, Alistair and Lungren, Matt and Peng, Yifan and Lu, Zhiyong and Mark, Roger and Berkowitz, Seth and Horng, Steven},
  title = {{MIMIC-CXR-JPG - chest radiographs with structured labels}},
  journal = {{PhysioNet}},
  year = {2019},
  month = nov,
  note = {Version 2.0.0},
  doi = {10.13026/8360-t248}
}

@article{Nguyen2022,
  author    = {Ha Q. Nguyen and Khanh Lam and Linh T. Le and Hieu H. Pham and Dat Q. Tran and Dung B. Nguyen and Dung D. Le and Chi M. Pham and  Hang T. T. Tong and Diep H. Dinh and Cuong D. Do and Luu T. Doan and others},
  title     = {{VinDr-CXR: An open dataset of chest X-rays with radiologist’s annotations}},
  journal   = {Scientific Data},
  year      = {2022},
  volume    = {9},
  number    = {1},
  pages     = {429},
  doi       = {10.1038/s41597-022-01498-w},
  issn      = {2052-4463}
}

@INPROCEEDINGS{Cui2019,
  author={Cui, Yin and Jia, Menglin and Lin, Tsung-Yi and Song, Yang and Belongie, Serge},
  booktitle={2019 IEEE/CVF Conference on Computer Vision and Pattern Recognition (CVPR)}, 
  title={Class-Balanced Loss Based on Effective Number of Samples}, 
  year={2019},
  volume={},
  number={},
  pages={9260-9269},
  doi={10.1109/CVPR.2019.00949}}

@INPROCEEDINGS{Tal2021,
  author={Ridnik, Tal and Ben-Baruch, Emanuel and Zamir, Nadav and Noy, Asaf and Friedman, Itamar and Protter, Matan and Zelnik-Manor, Lihi},
  booktitle={2021 IEEE/CVF International Conference on Computer Vision (ICCV)}, 
  title={Asymmetric Loss For Multi-Label Classification}, 
  year={2021},
  volume={},
  number={},
  pages={82-91},
  doi={10.1109/ICCV48922.2021.00015}}

@inproceedings{hardt2016,
 author = {Hardt, Moritz and Price, Eric and Srebro, Nati},
 booktitle = {Advances in Neural Information Processing Systems},
 editor = {D. Lee and M. Sugiyama and U. Luxburg and I. Guyon and R. Garnett},
 pages = {},
 publisher = {Curran Associates, Inc.},
 title = {Equality of Opportunity in Supervised Learning},
 volume = {29},
 year = {2016}
}

@article{LIN2025103739,
title = {CXR-LT 2024: A MICCAI challenge on long-tailed, multi-label, and zero-shot disease classification from chest X-ray},
journal = {Medical Image Analysis},
volume = {106},
pages = {103739},
year = {2025},
issn = {1361-8415},
doi = {10.1016/j.media.2025.103739},
author = {Mingquan Lin and Gregory Holste and Song Wang and Yiliang Zhou and Yishu Wei and Imon Banerjee and Pengyi Chen and Tianjie Dai and Yuexi Du and Nicha C. Dvornek and Yuyan Ge and Zuwei Guo and Shouhei Hanaoka and Dongkyun Kim and Pablo Messina and Yang Lu and Denis Parra and Donghyun Son and Álvaro Soto and Aisha Urooj and René Vidal and Yosuke Yamagishi and Pingkun Yan and Zefan Yang and Ruichi Zhang and Yang Zhou and Leo Anthony Celi and Ronald M. Summers and Zhiyong Lu and Hao Chen and Adam Flanders and George Shih and Zhangyang Wang and Yifan Peng}}

@misc{Dong2026,
  title        = {CXR-LT 2026 Challenge: Multi-Center Long-Tailed and Zero Shot Chest X-ray Classification},
  author       = {Dong, Hexin and Lin, Yi and Zhou, Pengyu and Zhao, Fengnian and Legasto, Alan Clint and Cho, Juno and Kim, Dohui and Kim, Justin Namuk and Kim, Mingeon and Kwak, Sunwoo and others},
  year         = {2026},
  eprint       = {2604.15555},
  archivePrefix= {arXiv},
  primaryClass = {eess.IV}
}

@article{Sounderajah2025,
  author  = {Sounderajah, Viknesh and
             Guni, Ahmad and
             Liu, Xiaoxuan and
             Collins, Gary S. and
             Karthikesalingam, Alan and
             others},
  title   = {The STARD-AI reporting guideline for diagnostic accuracy studies using artificial intelligence},
  journal = {Nature Medicine},
  year    = {2025},
  volume  = {31},
  number  = {10},
  pages   = {3283--3289},
  doi     = {10.1038/s41591-025-03953-8},
  issn    = {1546-170X}
}

@article{sagawa2020,
  title={Distributionally robust neural networks for group shifts: On the importance of regularization for worst-case generalization},
  author={Sagawa, Shiori and Koh, Pang Wei and Hashimoto, Tatsunori B and Liang, Percy},
  journal={arXiv preprint arXiv:1911.08731},
  year={2019}
}

@article{Gianfrancesco2020,
  author    = {Gianfrancesco, Milena A. and Tamang, Suzanne and Yazdany, Jinoos and Schmajuk, Gabriela},
  title     = {Potential Biases in Machine Learning Algorithms Using Electronic Health Record Data},
  journal   = {JAMA Internal Medicine},
  year      = {2018},
  month     = nov,
  volume    = {178},
  number    = {11},
  pages     = {1544--1547},
  doi       = {10.1001/jamainternmed.2018.3763},
  pmid      = {30128552},
  pmcid     = {PMC6347576},
  issn      = {2168-6106}
}

@article{Larrazabal2020,
  author  = {Larrazabal, Agostina J. and Nieto, Nicol{\'a}s and Peterson, Victoria and Milone, Diego H. and Ferrante, Enzo},
  title   = {Gender imbalance in medical imaging datasets produces biased classifiers for computer-aided diagnosis},
  journal = {Proceedings of the National Academy of Sciences of the United States of America},
  year    = {2020},
  month   = jun,
  volume  = {117},
  number  = {23},
  pages   = {12592--12594},
  doi     = {10.1073/pnas.1919012117},
  pmid    = {32457147},
  pmcid   = {PMC7293650},
  issn    = {0027-8424}
}

@article{SeyyedKalantari2021,
  author  = {Seyyed-Kalantari, Laleh and Zhang, Haoran and McDermott, Matthew B. A. and Chen, Irene Y. and Ghassemi, Marzyeh},
  title   = {Underdiagnosis Bias of Artificial Intelligence Algorithms Applied to Chest Radiographs in Under-Served Patient Populations},
  journal = {Nature Medicine},
  year    = {2021},
  volume  = {27},
  number  = {12},
  pages   = {2176--2182},
  doi     = {10.1038/s41591-021-01595-0},
  issn    = {1546-170X}
}

@inproceedings{
Yang2024,
title={A Textbook Remedy for Domain Shifts: Knowledge Priors for Medical Image Analysis},
author={Yue Yang and Mona Gandhi and Yufei Wang and Yifan Wu and Michael S Yao and Chris Callison-Burch and James Gee and Mark Yatskar},
booktitle={The Thirty-eighth Annual Conference on Neural Information Processing Systems},
year={2024}
}

@article{Johnson2023,
  author  = {Johnson, Alistair E. W. and Bulgarelli, Lucas and Shen, Lu and Gayles, Alvin and Shammout, Ayad and Horng, Steven and Pollard, Tom J. and Hao, Sicheng and Moody, Benjamin and Gow, Brian and Lehman, Li-wei H. and Celi, Leo A. and Mark, Roger G.},
  title   = {MIMIC-IV, a Freely Accessible Electronic Health Record Dataset},
  journal = {Scientific Data},
  year    = {2023},
  volume  = {10},
  number  = {1},
  pages   = {1},
  doi     = {10.1038/s41597-022-01899-x},
  url     = {https://doi.org/10.1038/s41597-022-01899-x},
  issn    = {2052-4463}
}

\end{document}